\documentclass[runningheads]{llncs}
\usepackage{graphicx}
\usepackage[misc,geometry]{ifsym} 
\usepackage{latexsym}
\usepackage{comment}
\usepackage{multirow}
\usepackage{subfig}
\usepackage{amssymb}
\usepackage{amsmath}
\usepackage{array}
\usepackage{tabularx}
\usepackage[linesnumbered,algoruled,boxed,lined]{algorithm2e}
\SetKwInOut{Parameter}{parameter}
\begin{document}
\title{Automated Detection System for Adversarial Examples with High-Frequency Noises Sieve}
\titlerunning{Automated Detection System for Adversarial Examples}
%
\author{Dang Duy Thang (\Letter) \orcidID{0000-0003-1348-7252}\and
Toshihiro Matsui}
\authorrunning{Dang Duy Thang and Toshihiro Matsui}
%
\institute{The Institute of Information Security, Yokohama, Japan 
\\
\url{http://lab.iisec.ac.jp/~matsui\_lab/}\\
\email{\{dgs174101,matsui\}@iisec.ac.jp}}
\maketitle              
\begin{abstract}
Deep neural networks are being applied in many tasks with encouraging results, and have often reached human-level performance. However, deep neural networks are vulnerable to well-designed input samples called adversarial examples. In particular, neural networks tend to misclassify adversarial examples that are imperceptible to humans. This paper introduces a new detection system that automatically detects adversarial examples on deep neural networks. Our proposed system can mostly distinguish adversarial samples and benign images in an end-to-end manner without human intervention. We exploit the important role of the frequency domain in adversarial samples, and propose a method that detects malicious samples in observations. When evaluated on two standard benchmark datasets (MNIST and ImageNet), our method achieved an out-detection rate of 99.7--100\% in many settings.

\end{abstract}
\section{Introduction}\label{introduction}
Deep Neural Networks (DNNs) were developed as a machine learning approach to many complex tasks. Traditional machine learning methods are successful when the final value is a simple function of the input data. Conversely, DNNs can capture the composite relations between millions of pixels and textual descriptions, brand-related news, future stock prices, and other contextual information. DNNs attain state-of-the-art performance in practical tasks of many domains, such as natural language processing, image processing, and speech recognition~\cite{lecun2015deep}. Current state-of-the-art DNNs are usually designed to be robust to noisy data; that is, the estimated label of a DNN output is insensitive to small noises in the data. Noise robustness is a fundamental characteristic of DNN applications in real, uncontrolled, and possibly hostile environments. However, recent research has shown that DNNs are vulnerable to specially-crafted adversarial perturbations (also known as adversarial examples)~\cite{43405,Gu2014TowardsDN}, well-designed fluctuating inputs that are added to clean inputs. Developers of machine learning models assume a legitimate environment in both training and testing. Intuitively, the inputs \(X\) are assumed to come from the same distribution during both training and test times. That is, if the test inputs \(X\) are new and previously unseen during the training process, they at least have the same properties as the inputs used for training. These assumptions ensure a powerful machine learning model, but any attacker can alter the distribution during either the training time~\cite{ijcai2018-543} or the testing time~\cite{biggio2013evasion}. Typical training attacks~\cite{huang2011adversarial} try to inject adversarial training data into the original training set. If successful, these data will wrongly train the deep learning model. However, most of the existing adversarial methods attack the testing phase ~\cite{carlini2017,DBLP:conf/iclr/MadryMSTV18}, which is more reliable than attacking the training phase. Especially, training-phase attacks are more difficult to implement and should not be launched without first exploiting the machine learning system. For example, an attacker might slightly modify an image~\cite{ijcai2018-543}, causing it to be recognized incorrectly, or adjust the code of an executable file to enable its escape by a malware detector~\cite{grosse2017adversarial}. Many researchers have developed defense mechanisms against adversarial examples. For example, Papernot et al.~\cite{papernot2016distillation} deployed a distillation algorithm against adversarial perturbations. However, as pointed out by Carnili et al.~\cite{carlini2017}, this method cannot improve the robustness of a DNN system. Several other adversarial defense approaches have also been published~\cite{carlini2017,DBLP:conf/iclr/MadryMSTV18}. Carlini et al.~\cite{carlini2017} proposed new attacks based on three previously used distance metrics: $L_0, L_2$ and $L_{\infty}$, and evaluated the defenses of DNNs under the proposed attack methods. Madry et al.~\cite{DBLP:conf/iclr/MadryMSTV18} applied a natural saddle-point method that guards against adversarial examples in a principled manner. They found that the network architecture affects the adversarial robustness of a DNN, so the robust decision boundary of the saddle-point problem can be more complicated than a decision boundary that simply categorizes the legitimate data. Preprocessing-based defense strategies against adversarial examples, which are the focus of our current work, will be reviewed and discussed in Sec.~\ref{background}. 
\subsubsection{Our Contributions}
\begin{figure}[h]
	\centering
	\includegraphics[width=0.8\textwidth]{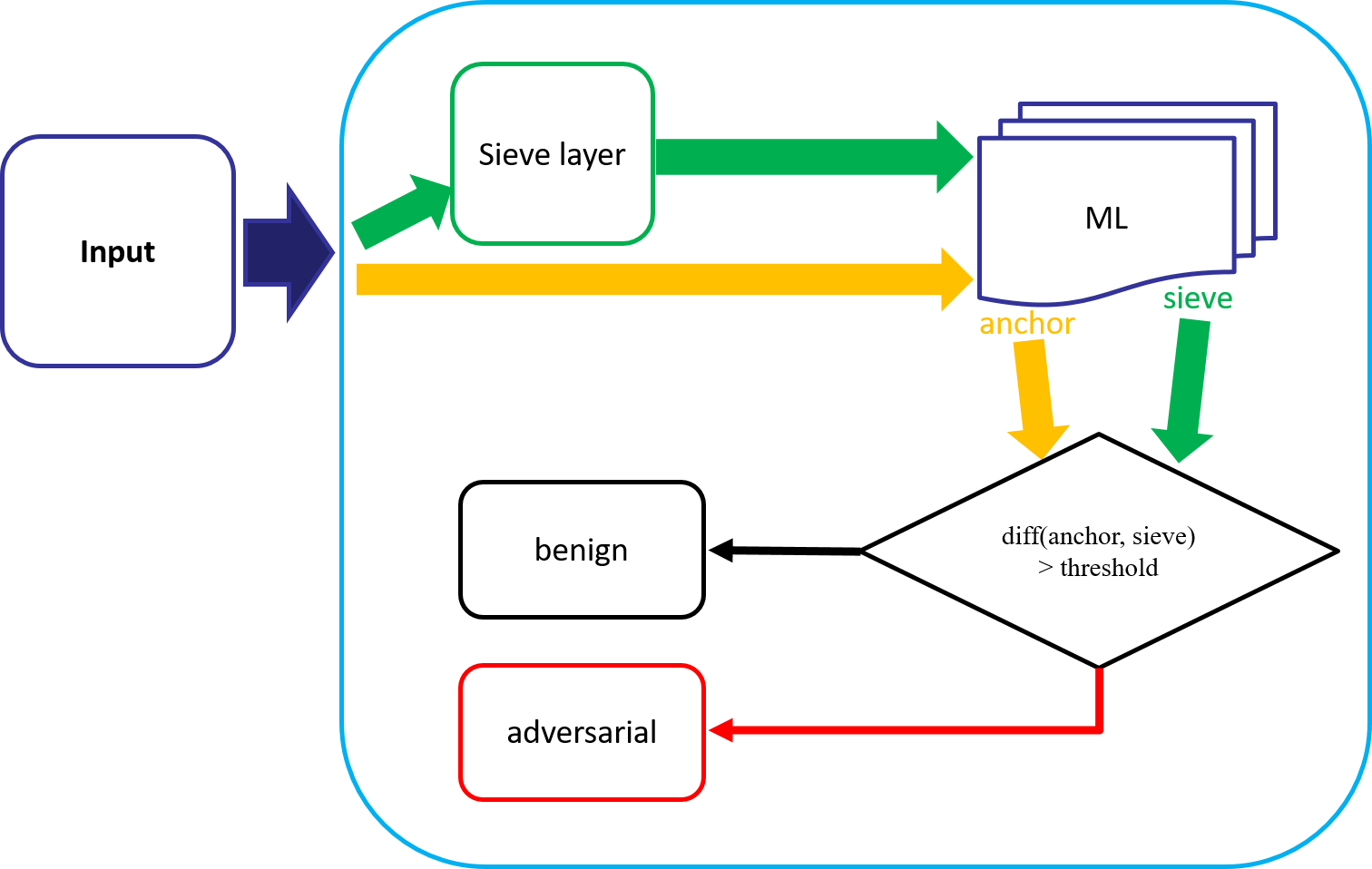}
	\caption{Processes of our automated detection system: an input is duplicated and the actual and duplicated inputs are directed into two parallel processing flows. In one processing direction (green arrow), the high-frequency adversarial noises are captured by the sieve layer; in the other direction (yellow arrow), all data are forwarded to the target model. The probability of the highest-confidence class is designated as the anchor for tracking in the green direction (called the sieve in our system). By comparing the difference between the anchor and sieve based on a fixed threshold, our system identifies the input as benign or adversarial.}
	\label{fig:our_system}
\end{figure}

This paper introduces new techniques for overcoming adversarial examples. Our proposed system can automatically detect and classify both adversarial and legitimate samples. Assuming that most of the adversarial perturbations are created in the high frequencies of the image, we seek to reduce the high-frequency adversarial noises while retaining the benign high-frequency features. To prove our hypothesis, we first installed a low-pass filter layer between the adversarial example and target classifier. The probability of detecting the target class by the classifier dropped significantly (to nearly zero), but the recognition results of the primary class were retained. In  Sec.~\ref{proposedmethod}, we demonstrate the correctness of these implementations in a theoretical proof. Based on the previous observation, we propose a new end-to-end system that automatically detects adversarial examples by a sieve layer inserted between the input and DNN, which traps suspicious noises. In parallel with this process, the un-sieved input is fed to the classifier and the highest-confidence class is marked as an anchor. The probabilities of the anchor and sieved input from the classifier are then compared, and the final decision on the input (adversarial or benign) is determined by a specified fixed threshold. Our main idea is depicted in Fig.~\ref{fig:our_system}.

The main contributions of this paper are as follows:
\begin{itemize}
\item We investigated and analyzed various attack approaches for crafting adversarial examples. By summarizing the different attack strategies, we provide an intuitive overview of these attack methods. 
\item We investigated the modern defense approaches and their variants in adversarial settings. We assumed that most of the adversarial perturbations are created at high frequencies. After implementing many experiments based on our theoretical framework, we confidently affirm our hypothesis. 
\item After thoroughly analyzing our experimental and theoretical observations, we created a new automated detection method for adversarial examples. Our approach differs from previous researches, in which the experimental steps are typically based only on the original hypothesis. Our approach was successfully applied to two types of common datasets: a small-scale dataset (Modified National Institute of Standards and Technology [MNIST]) and a large-scale dataset (ImageNet). Our defense method accurately classified the adversarial examples and legitimate samples. Moreover, in some cases, it recovered the high accuracy rates of the DNN classification.
\end{itemize}
\subsubsection{Outlines of the paper}
The remainder of this paper is organized as follows. Section \ref{background} introduces related works, the background of adversarial examples, and current adversarial attack/defense approaches. Section \ref{proposedmethod} describes our proposed method, and Section \ref{experiements} presents our experiments and evaluation results on both benchmark datasets (MNIST and ImageNet). Section \ref{conclusion} summaries our work.
\section{Related Works and Background}\label{background}

\subsection{Related Works} Removing the adversarial noises and regaining the recognition integrity of classifiers have been attempted in several works. Liao et al.~\cite{liao2018defense} developed High-level representation Guided Denoiser (HGD) as a defense for image classification systems. They argued that many defense models cannot remove all adversarial perturbations, and that the non-removed adversarial noises are greatly amplified in the top layers of the target model. Consequently, the model will output a wrong prediction. To overcome this problem, they trained a denoiser by an HGD loss function. However, their proposed system was implemented only on ImageNet, which contains color images, and was not trialed on grayscale datasets such as MNIST. Although this omission is not highly important, the performance of a method based on high-level representation in a very deep neural network may degrade on grayscale images, whereas a simple neural network performs accurately on MNIST data. The strategy of Xu et al.~\cite{DBLP:conf/ndss/Xu0Q18}, which they called "feature squeezing", reduces the number of degrees of freedom available to an adversary by squeezing out the unnecessary input features. The squeezing is performed by two denoisers performing different denoising methods: (1) squeezing the color bit depth, and (2) spatial smoothing. The prediction results are then compared with those of the target model, and the input is inferred as adversarial or legitimate. Although Xu et al.~\cite{DBLP:conf/ndss/Xu0Q18} evaluated their proposed method on various adversarial attacks, how they specified their thresholds on different benchmark datasets is unclear. Deciding appropriate thresholds for their system will overburden operators, and the method cannot easily adapt to new and unknown datasets.
\subsection{Background}
\subsubsection{Deep Neural Networks}
In this subsection, we review neural networks in detail and introduce the required notation and definitions. Neural networks consist of elementary computing units named neurons organized in interconnected layers. Each neuron applies an activation function to its input, and produces an output. Starting with the input to the machine learning model, the output produced by each layer of the network provides the input to the next layer. Networks with a single intermediate hidden layer are called shallow neural networks, whereas those with multiple hidden layers are DNNs. The multiple hidden layers hierarchically extract representations from the model input, eventually producing a representation for solving the machine learning task and outputting a prediction. A neural network model \(F\) can be formalized as multidimensional and parametrized functions $f_i$, each corresponding to one layer of the network architecture and one representation of the input. Specifically, each vector \(\theta_i\) parametrizes layer \(i\) of the network \(F\) and includes weights for the links connecting layer \(i\) to layer $i_1$. The set of model parameters $\theta = \{\theta_i\}$ is learned during training. For instance, in supervised learning, the parameter values are learned by computing the prediction errors $f(x)-y$ on a collection of known input--output pairs $(x, y)$.
\subsubsection{Adversarial Attacks}
The adversarial examples and their counterparts are considered to be indistinguishable by humans. Because human perception is difficult to model, it is often approximated by three distance metrics based on the \(L_p\) norm:
\begin{equation}
||x||_{p}=\bigg(\sum_{i=1}^n|x_{i}|^p\bigg)^\frac{1}{p}.
\end{equation}

The \(L_0, L_2, L_\infty\) metrics are usually used for expressing different aspects of visual significance. \(L_0\) counts the number of pixels with different values at the corresponding positions in two images. This measure describes the number of pixels that differ between two images. \(L_2\) measures the Euclidean distance between two images, and \(L_\infty\) helps to measure the maximum difference among all pixels at the corresponding positions in two images. The best distance metric depends on the proposed algorithms.

Szegedy et al.~\cite{szegedy2013intriguing} created targeted adversarial examples by a method called Limited-memory Broyden-Fletcher-Goldfarb-Shanno (L-BFGS), which minimizes the weighted sum of the perturbation size \(\varepsilon \) and loss function \(L(x^*,y_{target})\) while constraining the elements of \(x^*\) to normal pixel values.

According to Goodfellow et al.~\cite{43405}, adversarial examples can be caused by the cumulative effects of high-dimensional model weights. They proposed a simple attack method called the Fast Gradient Sign Method (FGSM):
\begin{equation}\label{eq:fgsm}
x^*= x+\varepsilon \cdot sign(\triangledown_xL(x,y)),
\end{equation}
where \(\varepsilon \) denotes the perturbation size for crafting an adversarial example \(x^*\) from an original input \(x\). Given a clean image \(x\), this method attempts to create a similar image \(x^*\) in the \(L_\infty\) neighborhood of \(x\) that fools the target classifier. This process maximizes the loss function \(L(x,y)\), which defines the cost of classifying image \(x\) as the target label \(y\). The FGSM solves this problem by performing one-step gradient updates from \(x\) in the input space with a small perturbation \(\varepsilon \). Increasing \(\varepsilon \) increases the magnitude and speed of the attack-success rate, but widens the difference between the adversarial sample and the original input. FGSM computes the gradients only once, so is much more efficient than L-BFGS. Despite its simplicity, FGSM is a fast and powerful generator of adversarial examples. FGSM maximizes the loss function \(L(x,y)\) by gradient descent (GD), a standard method for solving unconstrained optimization problems. Meanwhile, constrained problems can be solved by projected gradient descent (PGD). Madry el al.~\cite{DBLP:conf/iclr/MadryMSTV18} applied PGD in a new adversarial attack method defined as: 
\begin{flalign}\label{eq:pgd}
x^*=  x+\delta \cdot \left ( \bigtriangledown L\left ( x,y \right ) \right ) \textrm{ respect to project}  _{\left (x,\epsilon  \right ) }(x^*)&&
\end{flalign}
Where project$_{\left (x,\epsilon  \right )}$($x^*$) defines a projection operator with parameter $x^*$ on the circle area around $x$ with radius $\epsilon$, $\delta$ is a clip value that is searched in a box $(x,\epsilon)$. 
In this paper, we employ both FGSM and PGD in the attack phase. The adversarial attacks are created by a method called the attacking model. When the attacking model is the target model itself or contains the target model, the resulting attacks are white-box. In the present work, our method also operates in a white-box manner. 
\subsubsection{Adversarial Defenses}
Adversarial training of machine learning systems has been extensively researched~\cite{kurakin2016adversarial1,tramer2017ensemble}. This strategy trains the models on adversarial examples to improve their attack robustness. Some researchers have combined data augmentation with adversarial perturbed data for training~\cite{szegedy2013intriguing,kurakin2016adversarial1,tramer2017ensemble}. However, this training is more time consuming than traditional training on clean images alone, because it adds an extra training dataset to the training set, which clearly extends the training time. In other defense strategies based on pre-processing, the perturbation noise is removed before feeding the data into a machine learning model. Meng el al.~\cite{meng2017magnet} proposed a two-phase defense model that first detects the adversarial input, and then reforms the original input based on the difference between the manifolds of the original and adversarial examples. Another adversarial defense direction is based on the gradient masking method~\cite{tramer2017ensemble}. By virtue of the gradient masking, this defense strategy typically ensures high smoothness in specific directions and neighborhoods of the training data, inhibiting attackers from finding the gradients of the good candidate directions. Accordingly, they cannot perturb the input in a damaging way. Papernot et al.~\cite{papernot2016distillation} adapted distillation to adversarial defense, and trained the target model on soft labels output by another machine learning model. Nayebi et al.~\cite{nayebi2017biologically} conferred robustness to adversarial examples by saturating the network. The loss function in this strategy encourages the saturating regime of the activations. Gu et al.~\cite{Gu2014TowardsDN} proposed the deep contrastive network, which imposes a layer-wise contrastive penalty to achieve output invariance under input perturbations. However, methods based on gradient masking can be replaced by a substitute model (a copy that imitates the defended model), which attackers can train by observing the labels assigned by the defended model to inputs that are chosen carefully by the adversary.
\section{Proposed End-to-End System} \label{proposedmethod}
\subsection{Attack phase}
Our targeted attack settings are white-box, meaning that attackers can fully access the model type, model architecture, and all trainable parameters. The adversary aims to change the classifier's prediction to some specific target class. Using the available information, the attackers identify the vulnerable portion of the feature space, or seek the victim decision boundaries. The clean input of the victim model is then altered by adversarial example methods. Adversarial samples that will be misclassified by machine learning models are generated by an adversary with knowledge of the model's classifier $f$ and its trainable parameters. In this work, adversarial examples are created by the FGSM~\cite{43405} and PGD methods~\cite{DBLP:conf/iclr/MadryMSTV18}. We first define a classifier function \(f: \mathbb{R}^n \rightarrow   \begin{bmatrix}1...k\end{bmatrix} \) that maps the image pixel-value vectors to a particular label. We then assume a loss function \(L: \mathbb{R}^n \times  \begin{bmatrix}1...k\end{bmatrix}  \rightarrow  \mathbb{R} \) for function \(f\). Given an input image \(x \in  \mathbb{R}^n\) and target label \(y \in  \begin{bmatrix}1...k\end{bmatrix} \), our system attempts to optimize \(\delta+L(x+\delta,y)\) subject to \(x+\delta \in  \begin{bmatrix}0,1\end{bmatrix}^n \) , where \(\delta\) is a perturbation noise added to the original image \(x\). Note that this function solves \(f(x)\) in the case of convex losses, but can only approximately solve neural network problems, which involve non-convex losses. In this case, the gradient is computed not from the Softmax output, but from the output of the second-to-last layer logits.  Our PGD-based attack phase is described by Algorithm~\ref{alg:attack_phase}. In FGSM, Algorithm~\ref{alg:attack_phase} is executed without constraining $||\delta _{x} ||_{\infty}$. In the attack phase, the learning rate for crafting adversarial examples was set to 0.01 and the process was iterated 500 times. The targeted output images were created from clean input images.
\begin{algorithm}[h]
	\SetKwData{Left}{left}
	\SetKwData{This}{this}
	\SetKwData{Up}{up}
	\SetKwFunction{Union}{Union}
	\SetKwFunction{FindCompress}{FindCompress}
	\SetKwInOut{Input}{input}
	\SetKwInOut{Output}{output}
	\Input{$x$, $y_{true}$, $y^*$, $f$, $\epsilon$, $\alpha$}
	\Output{$x^*$}
	\Parameter{learning rate = 0.01, epochs = 500}
	\BlankLine
	$x$ $\leftarrow$ $x^*$	\tcp{initial adversarial sample}
	$\delta _{x} \leftarrow \vec{0}$ \tcp{initial perturbation factor}
	\While{$||\delta _{x} ||_{\infty}$ $< \epsilon$ and $f(x^*)\neq y^*$}{
		$x^* \leftarrow x+\alpha \cdot sign(\bigtriangledown L(y^*|x^*))$\\
		maximize $L(y^*|x^*)$ with respect to $||\delta _{x} ||_{\infty}$\\
		$x^* \leftarrow clip(x^*,x-\alpha,x+\alpha)$\\
		$\delta _{x} \leftarrow (x^*-x) $
	}
	\KwRet{$x^*$}
	\caption{Algorithm for Crafting Adversarial Examples }\label{alg:attack_phase}
\end{algorithm}
\subsection{Detection phase}
To create a new benchmark dataset for our detection system, we combined benign images with the adversarial images created in the attack phase. Assuming that the adversarial noises are high- frequency features on the images, we targeted the high-frequency domains on the images while retaining all features in the low-frequency areas. Various common algorithms are available for reducing image noises before further processing such as classification. In this work, we investigate the two most well-known filters in image denoising studies: linear and non-linear filters. For example, consider a new array with the same dimensions as the specified image. Fill each location of this new array with the weighted sum of the pixel values from the locations surrounding the corresponding location in the image, using a constant weight set. The result of this procedure is shift-invariant meaning that the output value depends on the pattern (not the positions) of the image neighborhood. It is also linear, meaning that summing the two images yields the same output as summing the separate outputs of both images.
This procedure, known as linear filtering, smooths the noises in the images. One famous linear filter is the Gaussian filter, defined as
\begin{equation}\label{eq:gaussian-filter}
G_\sigma (i,j)=\frac{1}{2\pi \sigma^2 }e^{-\frac{i^2+j^2}{2\sigma ^2} }.
\end{equation}
Here, \(i,j\) denotes the coordinate signal of the input and \(\sigma\) is the standard deviation of the Gaussian distribution. Alternatively, noise removal can be considered as filtering by a statistical estimator. In particular, the goal is to estimate the actual image value of a pixel in a noisy measurement scenario. The class of noise-removal filters is difficult to analyze, but is extremely useful. Smoothing an image by a symmetric Gaussian kernel replaces a pixel value with some weighted average of its neighbors. If an image has been corrupted by stationary additive zero-mean Gaussian noise, then this weighted average can reasonably estimate the original value of the pixel. The expected noise response is zero.  Weighting the spatial frequencies provides a better estimate than simply averaging the pixel values. However, when the image noise is not stationary additive Gaussian noise, the situation becomes more complicated. In particular, consider that a region of the image has a constant dark value with a single bright pixel composed of noise. After smoothing with a Gaussian, a smooth, Gaussian-like bright bump will be centered on the noise pixel. In this way, the weighted average can be arbitrarily and severely affected by very large noise values. The bump can be rendered arbitrarily bright by introducing an arbitrarily bright pixel, possibly by transient error in reading a memory element. When this undesirable property does not develop, the estimator outputs robust estimates. The most well-known robust estimator computes the median of a set of values from its neighborhood. A median filter assigns a neighborhood shape (which can significantly affect the behavior of the filter). As in convolution, this neighborhood shape is passed over the image, but the median filter replaces the current value of the element by median of the neighborhood values. For the neighborhood surrounding \((i\),\(j)\), the filter is described by:
\begin{equation}\label{eq:median-filter}
x_{ij}=median({X_{uv}|X_{uv}\in \mathbb{N}_{ij}}),
\end{equation}
where \(X_{uv}\) denotes the neighborhood points of \(x_{ij}\). Any adversarial noises can be attenuated by smoothing the pixels in the image. When adversarial noises are absent, smoothing the pixels does not severely affect the input-image quality, so the target classifier still recognizes the correct label. We name this process the sieve process (indicated by the green arrow in Fig.~\ref{fig:our_system}). 

Our proposed detection system runs the sieve and anchor processes in parallel. The sieve process arrests the high frequencies in the input processing while the anchor process transfers the input directly to the machine learning model. The probability of the highest-confidence class from the machine learning model is assigned as the anchor. The sieve process then tracks the oscillations of classes similar to the anchor class. If the probabilities \(p\) of the anchor and sieve differ by more or less than the fixed threshold \(\Theta\), our system confidently determines the input as adversarial or benign, respectively. Our system proceeds by Algorithm~\ref{alg:detection_system}, where \(\kappa\) denotes the kernel size, \(f\) is a machine learning function that computes the probabilities of the predicted class, and \(s\) is the sieve function. The sieve function based on the Gaussian filter is called the Detection System based on Gaussian (DSG); the other sieve function is Detection System based on Median (DSM).
\begin{algorithm}[h]
	\SetKwData{Left}{left}
	\SetKwData{This}{this}
	\SetKwData{Up}{up}
	\SetKwFunction{Union}{Union}
	\SetKwFunction{FindCompress}{FindCompress}
	\SetKwInOut{Input}{input}
	\SetKwInOut{Output}{output}
	\Input{$X, \Theta, s, f$}
	\Output{$0,1$} \tcp{0: benign; 1: adversarial}
	\Parameter{$\kappa=[(3\times3);(5 \times 5)]$}
	\BlankLine
	\For{$k_{size}$ in $\kappa $}{
		\For{$x$ in $X$}{
			$anchor_x$ $\leftarrow$ $x$\\
			$sieve_x$ $\leftarrow$ $x$\\
			$sieve_x$ $\leftarrow$ $s(sieve_x, k_{size})$\\
			$p(anchor_y)$ $\leftarrow$ $f(anchor_x)$\\
			$p(sieve_y)$ $\leftarrow$ $f(sieve_x|anchor_y)$\\
			
	}}
	\eIf{diff$(p(anchor_y), \textrm{min}(p(sieve_y))) > \Theta$}{
		\KwRet{$1$}
		
	}{
		\KwRet{$0$}
	}
	\caption{Automated Detection System of Adversarial Examples with a High-Frequency Sieve}\label{alg:detection_system}
\end{algorithm}
\section{Implementation and Results}\label{experiements}
\subsection{Datasets}
The classification task was evaluated on two common benchmark datasets, namely, MNIST and ImageNet. 
\subsubsection{Setup of MNIST}
The MNIST dataset~\cite{lecun2010mnist} includes 70,000 gray images of hand-written digits ranging from 0 to 9. It is separated into 60,000 training images and 10,000 testing images. A single MNIST image is composed of \(28\times28\) pixels, each encoded by an 8-bit grayscale. 
We randomly extracted 200 images of the digit ``0" from the 10,000 testing images. From each of these 200 images, we created nine adversarial images targeting the remaining digits (1-9). Finally we created a new benchmark dataset of 2,000 images (200 benign images and 1,800 adversarial images). 
\subsubsection{Setup of ImageNet}
The ImageNet dataset \cite{russakovsky2015imagenet} is a very large database designed for visual object recognition research. The original ImageNet includes more than 14 million images in 20,000 categories. Typical categories such as ``computer mouse\("\) and ``vending machine\("\) comprise several hundred images. As the machine learning model, we adopted Google Inception V3 \cite{szegedy2016rethinking} trained on 1,000 common categories in ImageNet.
We randomly selected 1,000 testing images (excluding ``ostrich" images). This selection of targeted class does not compromise the generality of our system. Applying the FGSM and PGD methods, we crafted 2,000 adversarial ``ostrich" images. Both image sets were combined into a new benchmark repository of 3,000 images for our experiment. 
\subsection{Implementation}
Although adversarial examples have recently attracted much interest from researchers, a public benchmark dataset for evaluating the robustness of defense systems remains lacking. In the attack phase of our system, we thus created a new benchmark dataset for evaluating the detection capabilities of our detection system. 

The 200 random images of digit ``0" extracted from the MNIST dataset were converted to adversarial images of digits 1-9 by the FGSM method. The FGSM was run through 1,000 iterations (epochs). The adversarial images were combined with original images into the new benchmark dataset for evaluating our detection system. The proposed detection system knows the true labels of the input. When presented with the unknown input, our system automatically processes the input and returns a decision (adversarial or benign). 

\begin{figure}[h]
	\centering
	\includegraphics[width=0.9\textwidth]{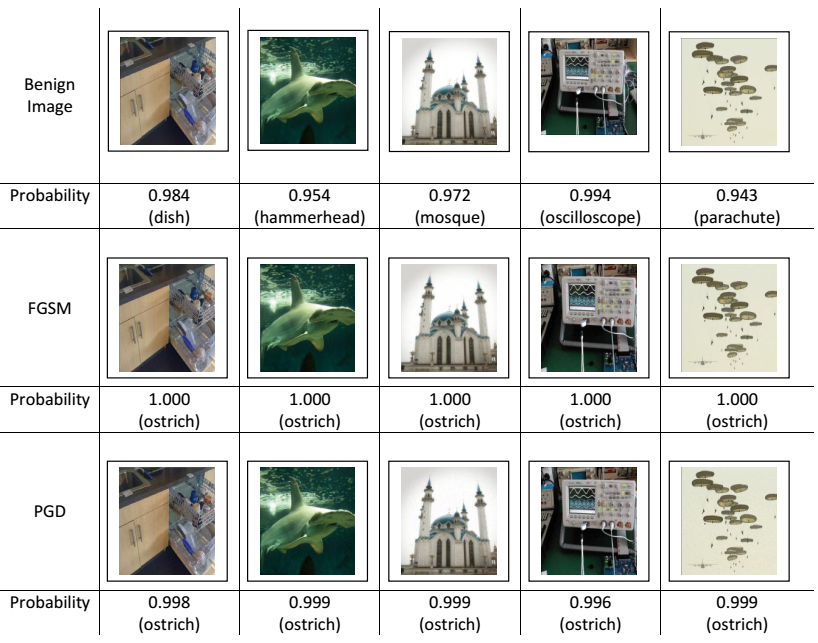}
	\caption{Attack Phase Samples.}
	\label{fig:imagenet-attack-phase}
\end{figure}

Applying FGSM and PGD, we created adversarial ``ostrich" images from 1,000 random ImageNet testing images. The original images were selected from categories other than ``ostrich". The FGSM and PGD were each run through 500 iterations (epochs). Figure~\ref{fig:imagenet-attack-phase} shows five samples among the 1,000 randomly picked images used for crafting the adversarial examples. The first row displays the original (benign) images of a dish, hammerhead, mosque, oscilloscope and parachute. Below the images, the probabilities present the highest probability of each image class (indicated in parentheses).

In the sieve process, the kernel sizes of the Gaussian and median filters were set to $(3\times3)$ and $(5\times5)$, respectively. Observing the probability changes of the anchor before and after sieving, the input data were assessed as benign or adversarial based on the given threshold.

\subsection{Results}
Our results were compared with those of Xu et al.~\cite{DBLP:conf/ndss/Xu0Q18}. Our system is more convenient that Xu's system, owing to its high detection accuracy and easy setup. Specifically, our system adopts a fixed threshold whereas Xu et al.'s system must adapt the threshold value to individual cases. The performance of our system was evaluated by the F1-score. When based on the Gaussian and median filtering, our detection system is called DSG and DSM, respectively. 

\begin{figure}[h]
	\centering
	\subfloat[Orginal Oscilloscope Image]{\label{figur:1}\includegraphics[width=35mm]{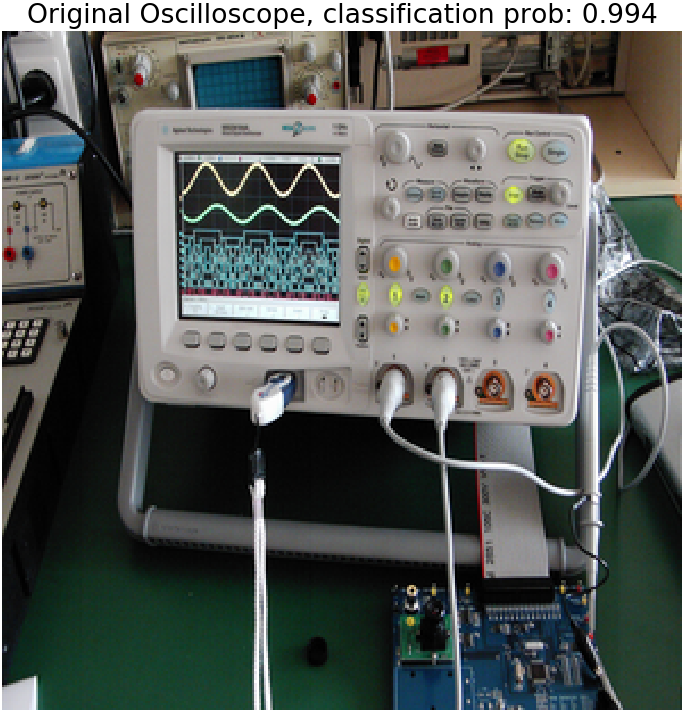}}
	\subfloat[Orginal Image with DSG (3x3)]{\label{figur:2}\includegraphics[width=35mm]{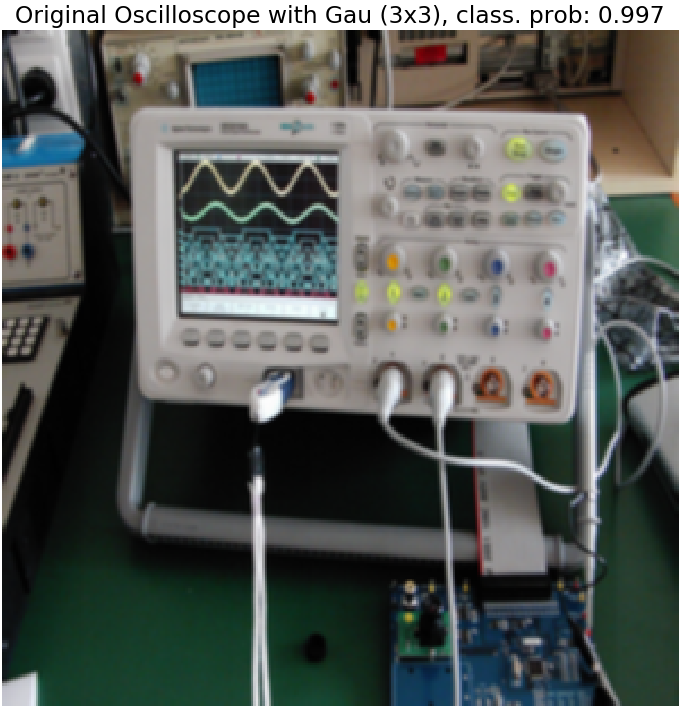}}
	\subfloat[Orginal Image with DSM (3x3)]{\label{figur:3}\includegraphics[width=35mm]{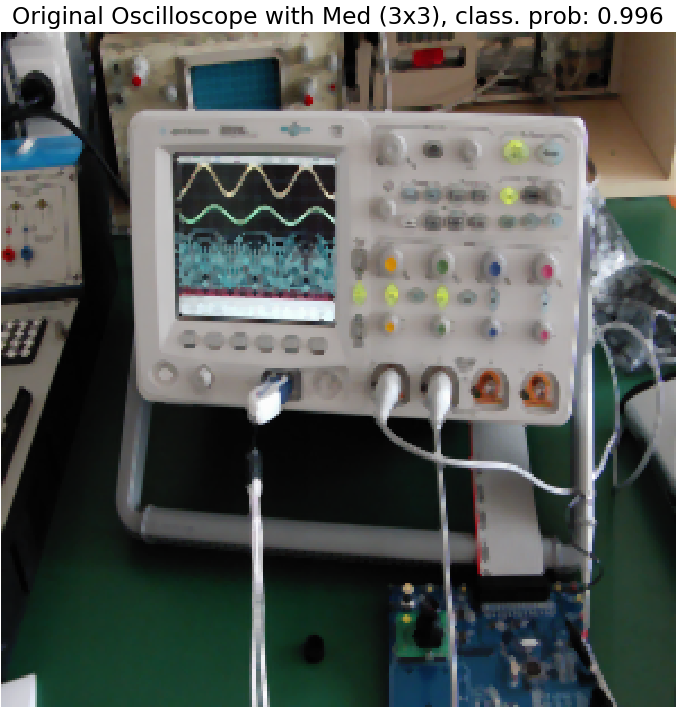}}
	\\
	\subfloat[Adversarial Image (Ostrich)]{\label{figur:4}\includegraphics[width=35mm]{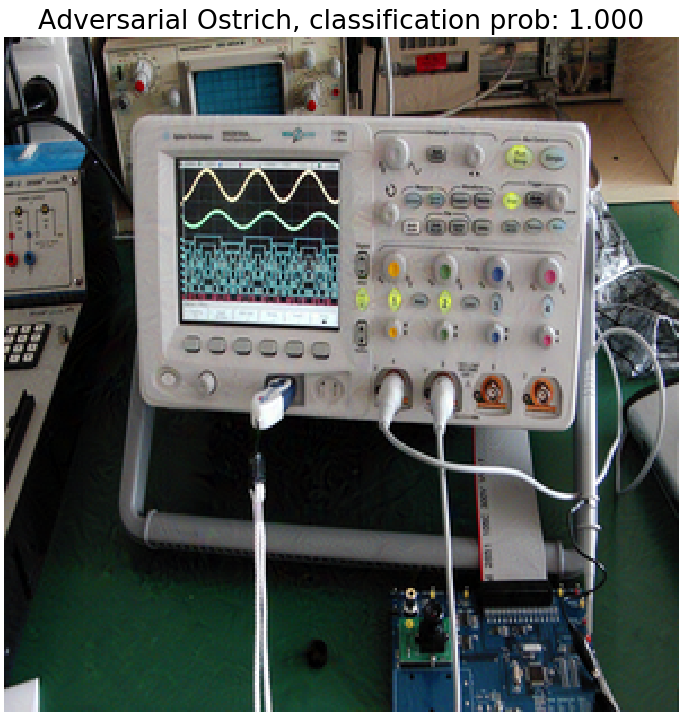}}
	\subfloat[Adversarial Image with DSG (3x3)]{\label{figur:5}\includegraphics[width=35mm]{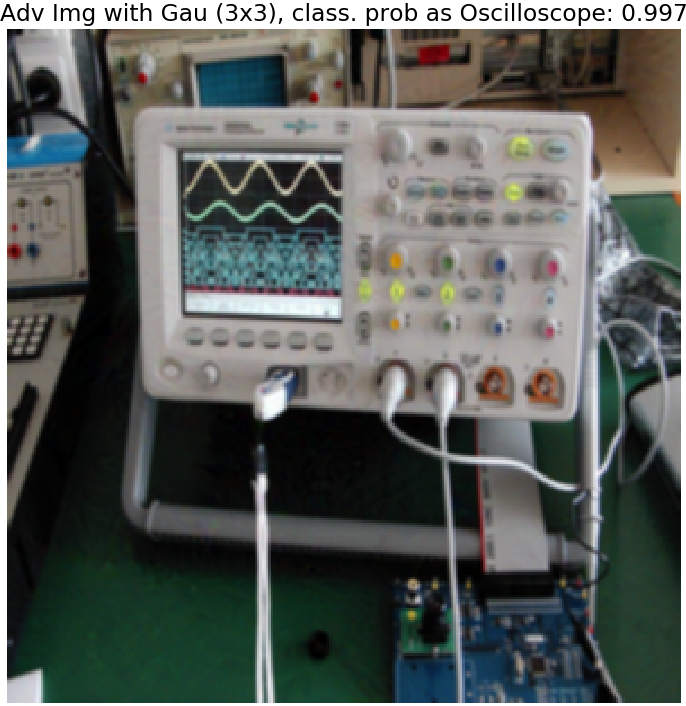}}
	\subfloat[Adversarial Image with DSM (3x3)]{\label{figur:6}\includegraphics[width=35mm]{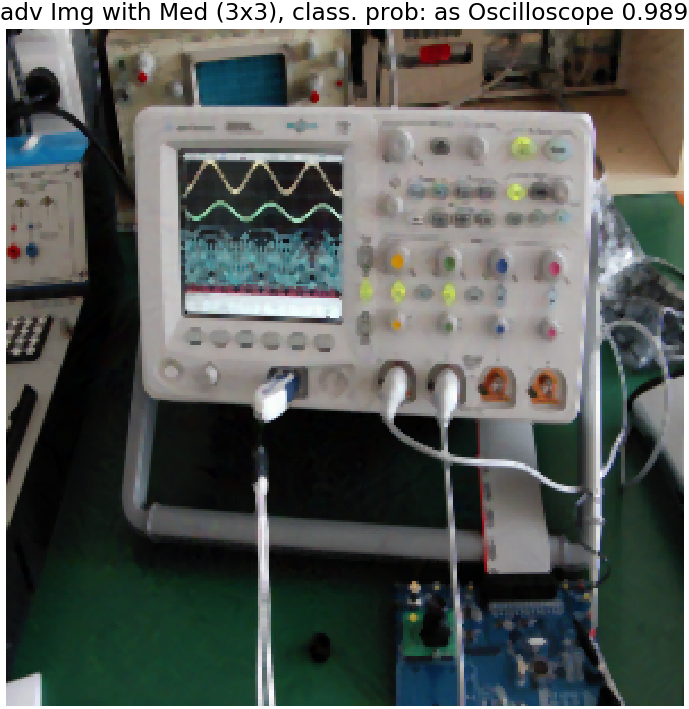}}
	\caption{Original image (true class: Oscilloscope) and Adversarial image suffer to DSG and DSM with kernel size is (3 $\times$ 3). (\ref{figur:1}) for original input, classification probability for oscilloscope label is 99.4\%, (\ref{figur:2}) for original input with DSG, classification probability for oscilloscope label is 99.7\%, (\ref{figur:3}) for original input with DSM, classification probability for oscilloscope label is 99.6\%, (\ref{figur:4}) for adversarial input, classification probability for ostrich label is 100\%, (\ref{figur:5}) for adversarial input with DSG, classification probability for oscilloscope label is 99.7\%, (\ref{figur:6}) for DSM input with classification probability for oscilloscope label is 98.9\%}%
	\label{fig:oscilloscope}%
\end{figure}

We observed and analyzed a typical oscilloscope image. From a benign oscilloscope image with a detection probability of 99.4\%, we created two adversarial images with the targeted label is ostrich, one by FGSM method, the other by PGD method (Fig.~\ref{fig:imagenet-attack-phase}). Afterward, the adversarial ostrich noises were sieved by the DSG and DSM functions, and the oscilloscope features were regained. As shown in Fig.~\ref{fig:oscilloscope}, the probabilities of the targeted ostrich and legitimate oscilloscope dramatically differed when processed by the DSG and DSM functions. When an input is Original Oscilloscope, classification probability is 99.4\% for oscilloscope label and the probabilities for true label are still remained round 99\% after using DSG or DSM algorithms. Conversely, with an adversarial image with targeted class ostrich, DSG and DSM not only remove adversarial noises but only regain the probabilities of true label nearly equal to when using original input. This observation confirms our assumption that adversarial noises are high-frequency noises, and that adversarial samples are powerfully detected by adopting the low-pass filter in our model.

The detection results on the MNIST dataset are reported in Table~\ref{tab:minist-detection}. The dashes in this table signify a lack of information from earlier research. Although the same number of images was compared in ours and Xu et al.'s methods, we created a more challenging test set than Xu et al.~\cite{DBLP:conf/ndss/Xu0Q18}. Whereas Xu et al. created a balanced dataset of 1,000 legitimate images and 1,000 adversarial examples, we created 1,800 adversarial images from 200 legitimate inputs, thus imposing an imbalanced~\cite{sun2009classification} dataset in our experimental test. Nevertheless, our detection rates are highly competitive with those of Xu et al. and slightly surpass the earlier detection rates. Moreover, our system applies a fixed threshold for all settings, whereas in Xu et al.'s work, the threshold must be adjusted in different settings.

\begin{table}
	\caption{Detection Rates on the MNIST dataset}
	\label{tab:minist-detection}
	\hbox to\hsize{\hfil
		\begin{tabular}{l|c|c|c|c|c}\hline\hline
			\multirow{2}{*}{}& \multicolumn{2}{c|}{Our Method} & \multicolumn{3}{c}{Xu et al.~\cite{DBLP:conf/ndss/Xu0Q18}} \\ \cline{2-6} 
			& DSG&DSM& Bit-Depth& Smoothing & Best-Joint \\ \hline \hline 
			No. Files   &  2,000  & 2,000  &  2,000& 2,000 & 2,000\\ 
			Threshold    & 0.1 & 0.1 &  0.0005 & 0.0029 & 0.0029 \\ 
			True Positive &  1799   & 1796 & -  & - & -  \\ 
			True Negative  &  198   & 195 & -  & - & -  \\
			False Positive &  2   & 5 & -  & - & -  \\
			False Negative &  1   & 4 & -  & - & -  \\
			Accuracy   &  0.999   & 0.996 & - & - & -  \\
			Precision   &  0.999   & 0.997 & -  & - & -  \\
			Recall   &  $\mathbf{0.999}$   & 0.998 & 0.903  & 0.868 & 0.982 \\
			F1 score   &  0.999   & 0.998 & -  & - & -  \\
			\hline
		\end{tabular}\hfil}
\end{table}

On the ImageNet dataset, our detection rates exceeded those of Xu et al. As highlighted in Table~\ref{tab:imagenet-detection}, we analyzed more files in this implementation than Xu et al., while maintaining the imbalance in our benchmark dataset. Our detection rate was 99.7\% amd 100\% with DSG and DSM, respectively, greatly outperforming Xu et al.'s system.

\begin{table}
	\caption{Detection Rates on the ImageNet dataset}
	\label{tab:imagenet-detection}
	\hbox to\hsize{\hfil
		\begin{tabular}{l|c|c|c|p{1.5cm}|c}\hline\hline
			\multirow{2}{*}{}& \multicolumn{2}{c|}{Our Method} & \multicolumn{3}{c}{Xu et al.~\cite{DBLP:conf/ndss/Xu0Q18}} \\ \cline{2-6} 
			& DSG&DSM& Bit-Depth& Smoothing & Best-Joint \\ \hline \hline 
			No. Files   &  3,000  & 3,000  &  1,800& 1,800 & 1,800\\ 
			Threshold    & 0.92 & 0.92 &  1.4417 & 1.1472 & 1.2128 \\ 
			True Positive &  1994   & 2000 & -  & - & -  \\ 
			True Negative  &  995   & 875 & -  & - & -  \\
			False Positive  &  45   & 125 & -  & - & -  \\
			False Negative &  6   & 0 & -  & - & -  \\
			Accuracy   &  0.983   & 0.958 & - & - & -  \\
			Precision   &  0.978   & 0.941 & -  & - & -  \\
			Recall   &  0.997   & $\mathbf{1.000}$ & 0.751  & 0.816 & 0.859  \\
			F1 score   &  0.987   & 0.999 & -  & - & -  \\
			\hline
		\end{tabular}\hfil}
\end{table}
\section{Conclusion}\label{conclusion}
We investigated the high-frequency noises in adversarial image examples. Based on the high-frequency noise assumption and a theoretical framework, we demonstrated the effectiveness of a low-pass filter in removing these noises. This observation guided the development of our automated detection system for adversarial examples. On the MNIST and ImageNet datasets, our system achieved maximum accuracy rates of $99.9\%$ and $100\%$, respectively. For evaluating our system, we constructed new benchmark datasets posing more challenges than previously constructed datasets~\cite{DBLP:conf/ndss/Xu0Q18,liao2018defense}. Whereas the earlier studies evaluated their systems on images from the training set, our evaluation employed the testing images. Although we also challenged our model on imbalanced datasets, our detection system delivered state-of-the-art performance. As another important contribution to the existing corpus, our system not only defeated adversarial noises, but also regained the legitimate class from adversarial examples. 
\section*{Acknowledgement}
	We would like to thank Professor Akira Otsuka for his helpful and valuable comments. This work is supported by Iwasaki Tomomi Scholarship.

%
%
%
\bibliographystyle{splncs04}
\bibliography{CSS_2019_China_revision}

\end{document}